\documentclass[11pt]{article}
\usepackage{paclic32}
\usepackage{times}
\usepackage{latexsym}

\usepackage{bm} 
\usepackage[pdftex]{graphicx} 
\usepackage{amsmath, amssymb} 
\usepackage{booktabs} 
\usepackage{amsmath } 
\usepackage{url}
\usepackage{amsmath}

\usepackage[whole]{bxcjkjatype}

\title{Long Short-Term Memory for Japanese Word Segmentation}

\author{
Yoshiaki Kitagawa\thanks{~~Now at Yahoo Japan Corporation.} \and Mamoru Komachi \\
Tokyo Metropolitan University\\
  6-6 Asahigaoka, Hino, Tokyo 191-0065, Japan\\
  {\tt ace1235813@gmail.com, komachi@tmu.ac.jp}
}

\date{}

\begin{document}
\maketitle

\begin{abstract}
This study presents a long short-term memory (LSTM) neural network approach to Japanese word segmentation (JWS). 
Previous studies on Chinese word segmentation have succeeded in using recurrent neural networks such as LSTM and gated recurrent units. 
However, in contrast to Chinese, Japanese includes several character types
    such as hiragana, katakana, and kanji, which produce orthographic
    variations and increase the difficulty of word segmentation. 
Additionally, while it is important to consider a global context,
traditional JWS approaches still rely on local features.
To address this problem, this study proposes employing an LSTM-based approach
to JWS.

\end{abstract}

\section{Introduction}
Word segmentation is a fundamental task of Japanese language processing. 
Moreover, word segmentation errors in East Asian languages (e.g., Japanese and
Chinese), which lack a trivial word segmentation process, can cause problems
for downstream NLP applications.
Thus, it is crucial to perform accurate word segmentation for the Japanese
language.

To achieve high accuracy, modern Japanese word segmentation (JWS) methods
utilize discriminative models relying on extensive feature engineering.
However, machine-learning-based methods tend to require hand-crafted feature
templates. Thus, they suffer from data sparseness.
Neural network models have, therefore, been investigated for various NLP tasks
to address the problem of feature engineering~\cite{liu2015multi,sutskever2014sequence,socher2013parsing,turian2010word,mikolov2013distributed}.
Neural network models enable the use of dense feature vectors (i.e.,
embeddings) that are learned via representation learning.

Another important problem in JWS corresponds to context modeling. Traditional
JWS methods employ feature templates to expand local features in a fixed
window. However, global information beyond the window is not considered.
Conversely, recurrent neural network (RNN) models grasp long distance
information owing to the use of long short--term memory (LSTM), achieving
state-of-the-art accuracy in Chinese word segmentation \cite{chen-EtAl:2015:EMNLP2}. 
However, it is uncertain whether the LSTM approach is also effective for JWS
because there are many types of character sets in Japanese that produce
orthographic variations.

Therefore, we propose an LSTM network for JWS that incorporates
character-level embeddings and long-distance dependency.
The main contributions of this study are as follows.

\begin{itemize}

\item We propose an LSTM model for JWS and investigate methods to utilize
    sparse features, such as character type, character $n$-gram, and dictionary
    features.

\item The experimental results indicate that the proposed word segmentation
    model achieves comparable performance to conventional approaches in both
    token- and sentence-level accuracy with respect to various datasets.

\item Our souce code is available at GitHub\footnote{\url{https://github.com/ace12358/WordSegmentation}}.

\end{itemize}

\section{LSTM for Japanese Word Segmentation}

\begin{figure*}[t]
\centering
\includegraphics[width=0.9\textwidth]{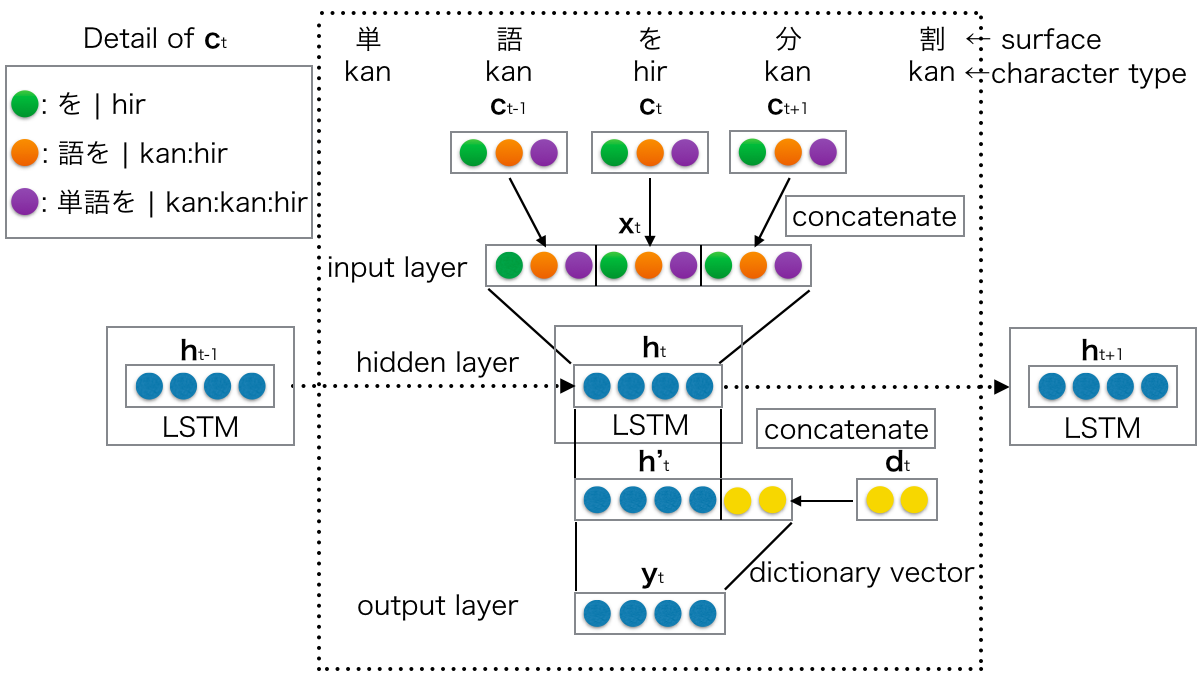}              
\caption{An overview of the proposed LSTM for JWS.}    
\label{ws_arch}    
\end{figure*}

Machine-learning-based approaches for word segmentation build a classifier
from an annotated corpus to classify the existence of word boundaries around a
target character.
In word segmentation, each character is assigned to several labels, such as
\{B, I, E, S\}, \{B, I, E\}, and \{B, I\} to indicate the segmentation, where
\{B\}, \{I\}, \{E\}, and \{S\} represents {\it Begin}, {\it Inside}, {\it
End}, and {\it Single}, in that order. 
In JWS, the most prevalent label set corresponds to \{B, I, E, S\},
and the label sets do not significantly affect the accuracy of our preliminary
experiments.

Classification of these labels is performed by running the Viterbi algorithm
over a word
lattice~\cite{kudo-yamamoto-matsumoto:2004:EMNLP,nakagawa2004chinese,kaji-kitsuregawa:2013:IJCNLP}
or by independently performing
predictions~\cite{neubig-nakata-mori:2011:ACL-HLT2011}. However, previous
approaches used feature templates to expand window-based local features.
Thus, they suffered data sparseness and a lack of global information in a
sentence.
An RNN, such as LSTM, addresses the problem of the lack of history by using
recurrent hidden units, in which the output at each time depends on that of the previous time. 
This method has been successfully demonstrated with respect to several NLP
tasks, such as language modeling~\cite{mikolov2010recurrent} and text
generation~\cite{sutskever2011generating}.

Thus, we propose character-based embeddings and an LSTM network for JWS.
Figure \ref{ws_arch} shows an overview of the proposed framework. The model is
similar to previous studies on CWS~\cite{chen-EtAl:2015:EMNLP2} which uses
character embeddings. However, our model also incorporates character-based
$n$-gram embeddings (character $n$-gram and character type $n$-gram) and a
word dictionary sparse feature in addition to character embeddings.

In the neural architecture, character-based embeddings for context characters
are extracted via the lookup table layer and concatenated into a single
vector,
$\bm{x}_t \in \mathbb{R}^{H_{1}}$, where $H_{1}$ is the size of the input layer. 
Thereafter, $\bm{x}_t$ is passed into the next layer to perform the linear
transformation, $\bm{W_{1}}$,  followed by an element-wise activation
function, $g$, such as sigmoid and $\tanh$ functions:
\begin{equation}
\label{ffnn_h}
 \bm{h}_t = g(\bm{W_{1}}\bm{x}_t + \bm{b_{1}})
\end{equation}
where $\bm{W_1} \in \mathbb{R}^{H_{2} \times H_{1}}$,
$\bm{b_{1}} \in \mathbb{R}^{H_{2}}$, and $\bm{h}_t \in \mathbb{R}^{H_{2}}$. 
Additionally, $H_{2}$ denotes a hyperparameter, which indicates the number of
hidden units in the hidden layer. $\bm{b_{1}}$ denotes a bias vector, and
$\bm{h}_t$ denotes the resulting hidden vector. The final output is obtained
by running a softmax function after a similar linear transformation,
$\bm{W_2}$, to the hidden vector as follows:
\begin{equation}
 \bm{y}_t = \mathit{softmax}(\bm{W_{2}}\bm{h}_t + \bm{b_{2}})
\end{equation}
where $\bm{W_2}$$\in$ $\mathbb{R}^{|T| \times H_{2}}$,
$\bm{b_{2}}\in\mathbb{R}^{|T|}$, and $\bm{y}_t \in \mathbb{R}^{|T|}$. Thus,
$\bm{b_{2}}$ denotes a bias vector, and $\bm{y}_t$  denotes the distribution
vector for each possible label.

\subsection{Character-Level Features}
This section discusses character-level features, as shown in Figure
\ref{ws_arch}. This paper introduces character embedding, character-type
embedding, and their $n$-gram for JWS. We describe the character vector,
$\bm{c}_t$, for JWS below. Formally, the character vector, $\bm{c}_t$,
is defined as follows:
\begin{equation}
\bm{c}_t = \bm{l}_t \oplus \bm{e}_t,
\end{equation}
where $\oplus$ denotes concatenation of the vectors, and $\bm{l}_t$ and
$\bm{e}_t$ denote character embeddings and character-type embeddings,
respectively. These embeddings are fed to the input layer.

In the following subsections, we discuss three features frequently used in
JWS, and we describe their realization as embeddings in the proposed
architecture.

\subsubsection{Character Embeddings}

In a word segmentation task, a character dictionary, $C$, of size $|C|$ is
often created. 
Traditional machine-learning approaches that use feature templates treat each
character independently as a one-hot vector.
However, it is natural for a neural network model to represent discrete data
as distributed vectors or
embeddings~\cite{Bengio:2003:NPL:944919.944966,collobert:2008}. Representation
learning is an actively studied topic in NLP because it overcomes the data
sparseness problem. Thus, the same practice is followed to represent each
character as a real-valued vector, $\bm{v_c}$ $\in$ $\mathbb{R}^{d}$, where $d$ is the dimensionality of the vector space. 
With respect to each character, the corresponding character embedding,
$\bm{v_c}$, is selected by a lookup table. 


\subsubsection{Character-Type Embeddings}
Character embeddings are extremely effective in identifying prefixes and
postfixes. However, they can be too sparse when crossing a word boundary.
To address this problem, it is helpful to exploit character types,
such as hiragana, katakana, and kanji (e.g.,
ひらがな，カタカナ，漢字), for JWS~\cite{neubig-nakata-mori:2011:ACL-HLT2011}.
For example, katakana sequences tends to correspond to a loan word. A
transition from a character type to another will likely correspond to a word
boundary~\cite{nagata1999part}.


\subsubsection{Character-Based $n$-gram Embeddings}
In addition to character type, the $n$-gram is effective in JWS \cite{neubig-nakata-mori:2011:ACL-HLT2011}.
Thus, the character-type sequence information is incorporated as embeddings. 
Each character is converted to a one-hot vector corresponding to its character
type. A one-hot vector comprises either hiragana, katakana, kanji,
alphabet, number, symbol, start symbol, or terminal symbol.
The advantages of a deep neural network include dealing with sparse vectors by
converting them to dense vectors.
This enables the utilization of a sparse feature, such as character trigram. 
Additionally, a character-based $n$-gram is effective for sentence similarity,
part-of-speech tagging \cite{wieting-EtAl:2016:EMNLP2016}, and for Japanese
morphological analysis~\cite{neubig-nakata-mori:2011:ACL-HLT2011}. Therefore,
$n$-gram is used for character and character-type embeddings.
More precisely, a one-hot vector is created for each unigram, bigram, and trigram.
Each embedding is selected by a lookup table as well as unigram embeddings.

The embedding vectors $\bm{l}_t$ and $\bm{e}_t$ are defined as follows:
\begin{eqnarray}
\bm{l}_t &=& \bm{l}_{[t-2:t]} \oplus \bm{l}_{[t-1:t]} \oplus \bm{l}_{[t]} \\
\bm{e}_t &=& \bm{e}_{[t-2:t]} \oplus \bm{e}_{[t-1:t]} \oplus \bm{e}_{[t]} 
\end{eqnarray}
where $\bm{l}_{[a:b]}$ denotes the embedding for the strings from a to b. The
same holds for $\bm{e}_t$.

\subsection{Incorporating Word Dictionary}

\begin{figure*}[t]    
 \centering
 \includegraphics[width=0.8\textwidth]{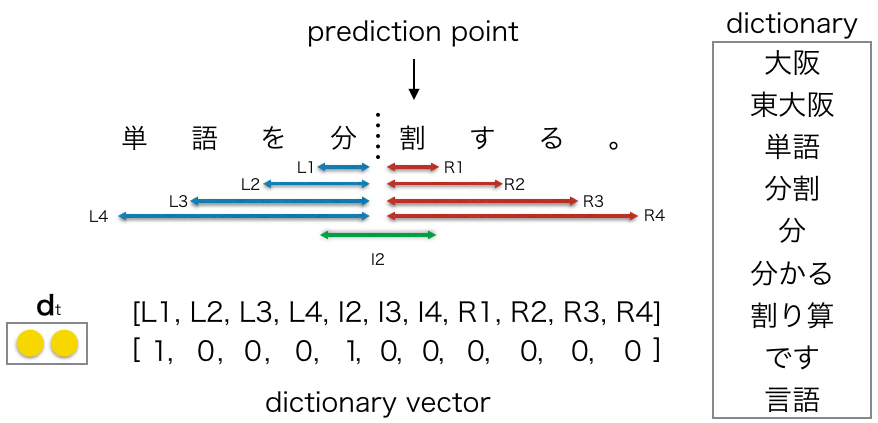}              
 \caption{Example of a dictionary vector.}    
 \label{dictionaryfeature}    
\end{figure*}

Character embeddings, character-type embeddings, and their $n$-gram extensions
perform an excellent job with respect to learning character-based features
from an annotated corpus. However, character-based JWS models lack word-level
information useful in determining the character sequences constituting a
word.
Thus, a Japanese morphological analyzer typically uses a dictionary.
It is essential for a JWS using a word lattice during decoding to use
word-level information such as a unigram and a bigram. However, this is not
necessary for character-based JWS approaches.

Notably, it is not trivial to encode dictionary information into a neural
network architecture.
\newcite{tsuboi:2014:EMNLP2014} suggests that it is ineffective to learn both
dense continuous and sparse discrete vector representations in the same layer.
Thus, we follow the same practice to create a sparse dictionary vector.
Whereas, instead of learning embeddings, this is used for the input to the
final output layer, as shown in Figure \ref{ws_arch}.

Figure \ref{dictionaryfeature} illustrates the creation of a dictionary
vector, comprising three parts, as follows: left-side feature $L$, right-side
feature $R$, and inside-feature $I$. For example, L2 is activated if a word
with a length corresponding to 2 exists in the dictionary on the left side of
the prediction point. If the length of the word exceeds a certain threshold,
the word length is cut off with respect to the length. In the study, 4 is
adopted as the threshold,
following~\newcite{neubig-nakata-mori:2011:ACL-HLT2011}.
In contrast to $L$ and $R$, 
$I2$ is fired if there exists a word spanning the boundary and possesses a
length of 2. It should be noted that $I$ is activated only if the length of
the word exceeds 1, based on its definition.
Finally, the feature vectors are concatenated to a single vector (e.g., a
dictionary vector).

The dictionary vector, $\bm{d}_t$, is concatenated to the current hidden
vector. It should be noted that the current hidden vector, $\bm{h}_t$, is on
top of the LSTM network.
Formally, the new hidden vector, $\bm{h^{\prime}}_t$, is defined as follows.
\begin{equation}  
\bm{h^{\prime}}_t = \bm{h}_t \oplus \bm{d}_t.
\end{equation}

\subsection{Training}
In this study, a cross-entropy error is adopted as a loss function. Given an
output vector, $\bm{y}_t$, the loss in a correct distribution corresponding to
$\bm{c}_t$ is computed as follows.
\begin{equation} 
 \bm{loss} = \sum_{t} -\bm{i}_t \log \bm{y}_t +  \frac{1}{2}\lambda
    \|\theta\|_{2}^2,
\end{equation}
where $\bm{i}_t$ denotes the correct label distribution, $\lambda$ denotes a
hyperparameter of L2 regularization, and $\theta$ indicates all parameters of
the model.
 
Following~\cite{socher2013parsing}, the diagonal variant of
AdaGrad~\cite{duchi2011adaptive} with mini batches is used to minimize the
objective. The update for the $i$-th parameter, $\theta_{t,i}$, at time step
$t$, is defined as follows.
\begin{equation}
\theta_{t,i} = \theta_{t-1,i} - \frac{\alpha }{\sqrt{\sum^{t}_{t=\tau}
    g^{2}_{\tau,i}}}g_{t,i}.
\end{equation}
where $\alpha$ denotes the initial learning rate, and $g_{\tau}$ denotes the gradient at time step $\tau$ for parameter $\theta_{i}$. 

\section{Experiments}

We evaluated the proposed neural word segmentation method on several JWS corpora.
To evaluate the neural network architectures, we prepare a feed-forward
network (FFNN) and an RNN for JWS. The FFNN is
illustrated by a dotted line in Figure \ref{ws_arch}. Additionally, the RNN
uses the same inputs as the LSTM, whereas it does not use any LSTM units.

The experiments are separated into two parts. First, the neural network
architectures and features are compared to previous state-of-the-art methods
on a balanced corpus. 
Second, the proposed method is evaluated on a newspaper corpus annotated with
a different segmentation criterion.

\subsection{Settings}
\begin{table}[t]
\caption{Number of sentences in the corpora.}
\label{bccwj}
\centering
\begin{tabular}{p{45mm}rr}
  \toprule
  \multicolumn{1}{c}{domain} & \multicolumn{1}{c}{train} &\multicolumn{1}{c}{test} \\
  \midrule
  Yahoo! Japan Answers  & 5,880 & 496 \\    
  Yahoo! Japan Blog     & 7,036 & 506 \\
  White paper           & 5,471 & 496 \\
  Magazine              &12,369 & 492 \\   
  Newspaper             &16,222 & 495 \\
  Book                  & 9,470 & 499 \\
  BCCWJ All             &56,448 &2,984\\
  \midrule
  KC All                &18,455 &1,234\\
  \bottomrule
\end{tabular}
\end{table}  

\paragraph{Datasets.}
We evaluate the methods with respect to two different datasets: a
popular Japanese corpus, the Balanced Corpus of Contemporary Written
Japanese (BCCWJ) version 1.1~\cite{maekawa2014balanced};
and another widely used Japanese corpus, the Kyoto University Corpus (KC),
version 4.0.
The BCCWJ is composed of various domains, whereas KC only includes the
newswire domain. The details of the corpora are shown in Table \ref{bccwj}.
The train and test split of BCCWJ follow, per the Project Next
NLP\footnote{\url{https://goo.gl/QCxxwB}}. We used a short unit word as
the segmentation standard, and we adopted the same train and test split of KC used in previous
studies~\cite{kudo-yamamoto-matsumoto:2004:EMNLP,uchimoto2001unknown}.

With respect to word-level features,
\newcite{neubig-nakata-mori:2011:ACL-HLT2011} do not use any external
dictionary, except the dictionary created from the training corpus. Hence, the
same scenario is adopted, and all the words in the training corpus are added.
However, words appearing only once in a corpus are omitted to prevent
overfitting of the training data, as
described in \cite{neubig-nakata-mori:2011:ACL-HLT2011}.
To analyze the effect of the dictionary feature, we recreate a larger
dictionary created from both training and test sets. This is termed as ``gold
dict'' in Table \ref{state_result}.

\paragraph{Tools.}
In the experiments, we use the state-of-the-art JWS tool, KyTea (ver.0.4.6)
\footnote{\url{http://www.phontron.com/kytea/}}, which implements
\cite{neubig-nakata-mori:2011:ACL-HLT2011} on this
dataset.\footnote{\newcite{morita-kawahara-kurohashi:2015:EMNLP}
used a different segmentation standard than ours, thus it is not directly
applicable to our dataset.}
We train a KyTea model using the provided scripts for training. This
internally creates a dictionary, as described above. Pretrained KyTea models
adopt their own word segmentation criterion, extended from that of BCCWJ.
Thus, KyTea models are retrained to ensure a fair comparison.

Additionally, we implement neural network-based JWS models, including FFNN,
RNN, and LSTM, by using a neural-network framework, Chainer (ver
1.4.0)\footnote{\url{http://chainer.org}}\cite{chainerlearningsys2015}.

\subsection{Hyperparameters}
We investigate several parameter combinations inspired by previous
studies~\cite{chen-EtAl:2015:EMNLP2} in our preliminary experiments. The
complete set of parameters used in the study is shown in Table
\ref{parameter}. The BCCWJ development set is used for tuning hyperparameters.

\paragraph{Pretraining.}
Based on the preliminary experiments and the early convergence of the learning
curve on the development set, we do not perform pretraining for character
embeddings.

\paragraph{Window size.} Preliminary experiments indicate that a window size
of 5 is better than others in terms of both accuracy and training time.
Thus, window size 5 is selected.

\paragraph{Dimension of character and character-type embeddings.}
The dimension of character embeddings is fixed by following \cite{chen-EtAl:2015:EMNLP2}.
In contrast, we search six configurations of character-type embeddings: 1, 3,
5, 10, 20, and 50. We set the hidden units of character-type embeddings to 10
because of the preliminary experiments.

\paragraph{Label set.}  In CWS, the label set \{B, I, E, S\} is often used. In
contrast, various label sets are adopted in JWS. We explore three label sets
and show that \{B, I, E, S\} is slightly better than the others.

\paragraph{Learning rate.} In this task, the learning rate largely affects
accuracy. A small learning rate (such as 0.01) degrades accuracy and
significantly affects learning time. Thus, a learning rate of 0.1 is
selected for all the experiments.

\begin{table}[t]
\caption{The hyperparameter set of the study.}
\label{parameter}
\centering
 \begin{tabular}{ll}
    \toprule
    hyperparameter            & value \\
    \midrule
    window size                & 5  \\
    character embeddings       & 100\\
    character type embeddings  & 10 \\
    hidden layer size          & 150\\
    label set                  & \{B, I, E, S\} \\
    learning rate              & 0.1\\
    coefficient of L2 regularization & 0.0001\\
   \bottomrule
\end{tabular}
\end{table}

\subsection{Results}
Table \ref{state_result} shows the experimental results for the BCCWJ Corpus.
In the KC, the LSTM+ctype+$n$-gram+dict$_{\mathit{sys}}$ model obtained
an F1 of 96.47, whereas the baseline KyTea 0.4.6 achieved an F1 of 96.21.
Our LSTM-based method outperformed the state-of-the-art method
\cite{neubig-nakata-mori:2011:ACL-HLT2011}. Table \ref{error} illustrates the
performance of the two methods per domain breakdown. The accuracy of the
proposed method, in terms of token-level F1 and sentence-level accuracy,
exceeds those of the others in four out of six domains, resulting in
improvements in the overall performance.  These four domains contain more
orthographic variants than the other two.

\section{Discussion}
\paragraph{Models.}
Table \ref{state_result} shows that LSTM is superior to FFNN and RNN by using
the same feature set (character embeddings only). It demonstrates the
effectiveness of modeling a context by LSTM.

\paragraph{Character-type embeddings.} Comparing {\it LSTM} with {\it LSTM +
ctype}, F1 improves by 0.25 points. The result shows that character-type
embeddings are useful in JWS.

\paragraph{Dictionary feature.} The addition of a dictionary feature to {\it
LSTM + ctype} improves F1 by 0.37. This result shows that dictionary feature
is effective in JWS. However, the addition of the dictionary feature to {\it
LSTM + ctype + $n$-gram} does not result in any notable difference. We
assume that character-based $n$-gram embeddings subsume the dictionary
feature because the dictionary is created from the training corpus,
(dict$_{\mathit{sys}}$). 
Additional experiments using the gold dictionary created from the test corpus,
(dict$_{\mathit{gold}}$), support this hypothesis\footnote{The singletons of
the combined corpus are removed while creating the gold dictionary. Thus the
test corpus may still contain words that are not in the gold dictionary.}.
Our findings are similar to \newcite{zhang-EtAl:2018:AAAI2018}, who
employed $n$-gram based feature templates and dictionaries for CWS.

\paragraph{$n$-gram embeddings.} A comparison of {\it LSTM + ctype} with {\it
LSTM + ctype + $n$-gram} indicates $n$-gram embeddings significantly improve
the performance of the model by a large margin. 
There are several attempts to incorporate neural representations into
a conditional random field (CRF) \cite{ma-hovy:2016:ACL2016,lemple-EtAl:2016:NAACL2016,peters-EtAl:2017:ACL2017},
all of which use bidirectional LSTM as encoders for sequence labeling tasks.
In contrast, we apply simple $n$-gram embeddings, which can be
easily obtained using a raw corpus. Our findings are in line with the rich
pretraining method for neural CWS \cite{yang-EtAl:2017:ACL2017}.

\begin{table}[t]
\caption{Experimental results of JWS on BCCWJ.
Ctype = character-type embeddings, and $n = 1,2,3$.}
\label{state_result}
\centering
\begin{tabular}{lcc}
\toprule
    Methods & F1 \\
  \midrule                  
      FFNN                              & 96.53 \\
      RNN                               & 96.46 \\
      LSTM                              & 97.00 \\
      LSTM+ctype                        & 97.25 \\    
      LSTM+ctype+dict$_{\mathit{sys}}$  & 97.37 \\
      LSTM+ctype+\{uni,bi\}gram         & 98.05 \\
      LSTM+ctype+$n$-gram               & 98.41 \\ 
      LSTM+ctype+$n$-gram+dict$_{\mathit{sys}}$  &{\bf 98.42} \\ 
      LSTM+ctype+$n$-gram+dict$_{\mathit{gold}}$ & 98.67 \\
  \midrule
      KyTea 0.4.6                       & 98.34 \\
  \bottomrule
\end{tabular}
\end{table} 

\section{Error Analysis}

\begin{table}[t]
\caption{Token-level and sentence-level performance on various domains in
    the BCCWJ dataset.}
\label{error}
\centering
\begin{tabular}{lrrrr}
  \toprule
  \multicolumn{1}{c}{Domain} & \multicolumn{2}{c}{F1} & \multicolumn{2}{c}{\# incorrect sent.} \\
  \multicolumn{1}{c}{} & KyTea & Ours & KyTea & Ours \\   
  \midrule
  Y! Answers    & 98.38       & {\bf 98.44} &75     & {\bf69}\\     
  Y! Blog       & {\bf 99.75} & 99.73       &98     & {\bf97}\\
  White paper   & {\bf 99.20} & 99.08       &{\bf81}& 84\\
  Book          & 98.15       & {\bf 98.28} &{\bf82}& 91\\ 
  Magazine      & 96.70       & {\bf 97.25} &102    & {\bf90} \\ 
  Newspaper     & 98.19       & {\bf 98.46} &96     & {\bf75}\\
  \midrule                                  
  All           & 98.34       & {\bf 98.42} &534    & {\bf506}\\
  \bottomrule
\end{tabular}
\end{table}

\begin{table}[t] 
\caption{An example of the error in this study and KyTea. The character ``\textbar '' indicates the word boundary, and the asterisk indicates the incorrect part.}
\label{error_analysis}
\centering
\begin{tabular}{ll}
  \toprule
  Method & Example\\
  \midrule
  Ours  & エルマー \textbar {\bf *とりゅう} \textbar の \textbar 絵 \textbar で \\
  KyTea & エルマー \textbar と \textbar りゅう \textbar の \textbar 絵 \textbar で \\ 
  \midrule
  Ours  & うち \textbar {\bf *がまんま} \textbar その \textbar 環境 \textbar です \textbar 。\\
  KyTea & うち \textbar が \textbar まんま \textbar その \textbar 環境 \textbar です \textbar 。\\
  \midrule
  Ours  & 七百 \textbar 六十 \textbar 一 \textbar の \textbar ため池 \textbar など\\  
  KyTea & 七百 \textbar 六十 \textbar 一
    \textbar {\bf *のため} \textbar 池 \textbar など \\
  \midrule
  Ours  & 思う \textbar と \textbar うんざり \textbar です \textbar ．\\
  KyTea & 思う \textbar {\bf *とうんざり} \textbar です \textbar ．\\
  \bottomrule
\end{tabular}
\end{table}

\subsection{Effect of Domain}
To determine the characteristics of the proposed method, we conducted
an error analysis by comparing the proposed method with KyTea, with respect to
different domains. Thus, we computed the F1 for each domain of BCCWJ, and
counted the number of incorrect sentences. Table \ref{error} summarizes
token-level and sentence-level comparisons between the proposed model and KyTea.

We selected Magazine that exhibited the largest margin in token-level F1 as
the successful domain, and selected Book and White paper having the largest
margin in sentence-level evaluation as the unsuccessful domains.

\paragraph{Magazine.} This domain contains colloquial expressions as well as
formal expressions. Hiragana occupies a substantial portion of this corpus
because of the colloquial expressions. Furthermore, F1 for this domain is the
lower in the two methods.
The results indicate that hiragana exhibits a poor performance. However, the
proposed method is more robust than KyTea in this domain. This may be caused
by the modeling of contextual information because the hiragana sequence tends
to fall outside of the local window size.

\paragraph{Book.}  This domain typically includes named entities, such as
a company name. This corpus is balanced in terms of the proportion of
character types.
Generally, the proposed model tends to be robust for compounds of different
character types (e.g., Famiポート (Fami Port)
multimedia vending machine), whereas
\newcite{neubig-nakata-mori:2011:ACL-HLT2011}'s model correctly classified
words comprising unique character types (e.g.ポストドクター (Postdoc)).
The difference between token-level and sentence-level accuracy
highlights the characteristic of these methods. The proposed method typically
produces fewer errors, whereas it does not consistently perform word
segmentation across the corpus.

\paragraph{White paper.} This domain comprises of official documents published
by the government. Thus, kanji covers a substantial portion of the corpus.
Additionally, the number of characters per sentence is high. In this domain,
the proposed method is only inferior to
\newcite{neubig-nakata-mori:2011:ACL-HLT2011}, with respect to both F1 and the
number of incorrect sentences. This is potentially caused by the long-sequence
introduced noise to the LSTM-based models.

\subsection{Example}
To investigate the characteristics of the proposed method from a
different perspective, we demonstrate actual examples of word segmentation.
Table \ref{error_analysis} shows a comparison of four examples for the current
study and KyTea 0.4.6.
The proposed method possesses two characteristics.

The first characteristic is that strings with the same character type tend to
form a word unit. This characteristic is demonstrated by the first and second
examples.
In the first example, \mbox{``と (and)''} and
\mbox{``りゅう (dragon)''} are different words.
However, they are of the same character type ``Hiragana.'' Thus, they are
incorrectly combined to form a fake word. In the second example,
\mbox{``が (NOM)'' } and \mbox{``まんま (just)''} are also incorrectly
connected. This type of error tends to occur when the character type
corresponds to \mbox{``Hiragana''}, which includes many high-frequency
ambiguous single-character particles.
 
Another characteristic of this method is that words with different
character types tend to be broken by KyTea at the position where a character
type is changed. This characteristic is demonstrated by the third example. In
this example, \mbox{``ため池 (storage reservoir)''} is a single word
consisting of \mbox{``ため (storage)''} and \mbox{``池 (reservoir)''},
whereas KyTea fails to recognize the word because \mbox{``ため''} and
\mbox{``池''} are of different character types. In contrast, the proposed
method correctly identifies the word.

However, there are cases where contrary results are indicated. In the fourth
example, \mbox{``と (and)''} and \mbox{``うんざり (fed up)''} correspond to
different words of the same character type \mbox{``Hiragana''}. An analysis of
the first and second examples indicates that the proposed method tends to form
a fake word that comprises of the same character type. However, it yields a
correct segmentation result. There is still room for improvement by using a
dictionary to address the problem of spurious words. The upper bound of
the proposed method is shown in Table \ref{state_result}.

\section{Related Works}
\label{sec:relatedworks}
In JWS, a supervised learning approach is widely used. A popular method in JWS
involves creating a word lattice by using a dictionary and using Viterbi
decoding~\cite{kudo-yamamoto-matsumoto:2004:EMNLP,sassano2002empirical}. This
approach is known to yield accurate results by considering the sequence of
words, whereas it is not robust if training data differ from test
data~\cite{neubig-nakata-mori:2011:ACL-HLT2011}.
Another popular approach employs pointwise prediction by using a local
window~\cite{neubig-nakata-mori:2011:ACL-HLT2011,NEUBIG10.408}. However, both
approaches do not consider the global context because they use feature
templates of a fixed length. Additionally, they both suffer from feature
sparseness.

Recently, deep neural network architectures have been widely used for CWS
tasks~\cite{chen-EtAl:2015:EMNLP2,chen-EtAl:2015:ACL-IJCNLP5,pei-ge-chang:2014:P14-1,zhang-zhang-fu:2016:P16-1,cai-zhao:2016:P16-1}.
These approaches are mainly divided into two types: structured prediction
model~\cite{zhang-zhang-fu:2016:P16-1,cai-zhao:2016:P16-1} and pointwise prediction model ~\cite{chen-EtAl:2015:EMNLP2,chen-EtAl:2015:ACL-IJCNLP5,pei-ge-chang:2014:P14-1}. However, a deep neural network approach requires high computational costs compared to previous approaches. 
In JWS, \newcite{morita-kawahara-kurohashi:2015:EMNLP} proposed integrating an
RNN language model into JWS by interpolating it with traditional JWS. As
opposed to using recurrent neural architecture as side information, word
segmentation in Japanese is directly learned by using LSTM.

Furthermore, a neural network approach for normalization was explored
\cite{kann-cotterell-schutze:2016:EMNLP2016,ikeda2016norm}.
\newcite{kann-cotterell-schutze:2016:EMNLP2016} proposed a character-based
encoder-decoder model and achieved state-of-the-art accuracy for the task of
canonical morphological segmentation.
Because their method was based on unsupervised learning, it could be learned
at a low cost. However, it was necessary to adjust word segmentation
criteria to human annotation.
\newcite{ikeda2016norm} also presented an encoder-decoder model for Japanese
text normalization. However, their model was only as good as conventional
CRF, although it was trained with a large-scale artificially created corpus.

\section{Conclusion}
In this paper, we presented an LSTM neural network approach to JWS.
We proposed learning Japanese-specific features, such as character-type and character
$n$-gram, as embeddings, and dictionary features as a sparse vector.
The proposed method was shown to achieve comparable accuracy to 
state-of-the-art systems on various domains. 

In JWS, it is important to deal with colloquial expressions that are
frequently found in dialogue-based conversations and web
text~\cite{saito-EtAl:2014:Coling,sasano-kurohashi-okumura:2013:IJCNLP,kaji-kitsuregawa:2014:EMNLP2014}.
It is expected that deep neural architectures, such as convolutional neural
networks, may be effective for this scenario because of their ability to learn
robust representations of characters and words~\cite{ling-EtAl:2015:EMNLP2}.

\section*{Acknowledgments}
This work was partly supported by the Microsoft Research Collaborative Research
(CORE) Projects. We thank anonymous reviewers for suggestions and comments,
which helped in improving the paper.

\bibliography{paclic32}
\bibliographystyle{acl}

\end{document}